\documentclass[preprint,12pt,authoryear]{elsarticle}

\usepackage{amssymb}
\usepackage[ruled, linesnumbered]{algorithm2e}
\usepackage{amsmath}
\usepackage{graphicx}
\usepackage{listings}
\usepackage{algpseudocode,algorithm2e}
\usepackage{array}
\usepackage{tabularray}
\usepackage{tabularx}
\usepackage{svg}
\usepackage{multirow}
\usepackage{url}
\usepackage{soul}
\usepackage{subfig}
\usepackage{rotating}
\usepackage{bm}

\def\ie{i.e.,\xspace}

\def\ourmodel{\textsc{IRECS}\xspace}

\journal{Expert Systems with Applications}

\begin{document}

\begin{frontmatter}



\title{Automatic selection of primary studies in systematic reviews with evolutionary rule-based classification}

\author[uco]{José de la Torre-López}\ead{i82toloj@uco.es}
\author[uco]{Aurora Ramírez}\ead{aramirez@uco.es}
\author[uco]{José Raúl Romero\corref{cor1}}\ead{jrromero@uco.es}
\affiliation[uco]{organization={Dept. of Computer Science and Artificial Intelligence},
            addressline={University of Córdoba}, 
            city={Córdoba},
            postcode={14071}, 
            state={AN},
            country={Spain}}
\cortext[cor1]{Corresponding author. email: \url{jrromero@uco.es}}

\begin{abstract}
Searching, filtering and analysing scientific literature are time-consuming tasks when performing a systematic literature review. With the rise of artificial intelligence, some steps in the review process are progressively being automated. In particular, machine learning for automatic paper selection can greatly reduce the effort required to identify relevant literature in scientific databases. We propose an evolutionary machine learning approach, called \ourmodel, to automatically determine whether a paper retrieved from a literature search process is relevant. \ourmodel builds an interpretable rule-based classifier using grammar-guided genetic programming. The use of a grammar to define the syntax and the structure of the rules allows \ourmodel to easily combine the usual textual information with other bibliometric data not considered by state-of-the-art methods. Our experiments demonstrate that it is possible to generate accurate classifiers without impairing interpretability and using configurable information sources not supported so far.
\end{abstract}


\begin{highlights}
\item An evolutionary machine learning approach to support literature selection in reviews.
\item Grammar-guided genetic programming to find classification rules for paper selection.
\item Knowledge from textual and bibliometric sources expressed through interpretable rules.
\item Experiments with different configurations show the contribution of bibliometric data.
\item Better performance compared to a black-box method based on active learning.
\end{highlights}


\end{frontmatter}



\section{Introduction}
\label{sec:intro}

Systematic Literature Reviews (SLR) are studies whose objective is to compile and analyse the most relevant and up-to-date research papers on a specific topic. Conducting a SLR is especially useful when starting a new line of research, as it involves a detailed analysis of the research topic supported by the appropriate references. This type of secondary study should be conducted following a strict protocol to ensure quality and allow replication~\citep{Booth2016book}.

Within the SLR process, manual and automated searches are performed to identify research papers related to the topic under review~\citep{Kichenham2007}. Therefore, the selection of primary studies, \ie papers of sufficient quality and truly relevant to the topic, is one of the most important steps. It is also a time-consuming task due to potentially large search results if the queries are too open-ended or the research topic is too broad. Recently, artificial intelligence (AI) has emerged as a way to assist researchers in this task, as well as in other stages of the SLR process~\citep{DeLaTorre2023}.
The topic has gained even more relevance since the appearance of Large Language Models (LLMs)~\citep{Han2024,Galli2025}. LLMs have expanded the capabilities of AI-assisted SLRs with the ability to extract information from papers, synthesise their findings and generate texts to accelerate SLR reporting. However, their use exposes researchers to new risks, such as lack of traceability, fictitious content generation, high computational cost, and biases~\citep{Han2024,Galli2025}.

Machine learning (ML) is especially suitable for primary study selection because of its ability to analyse large amounts of data and find patterns. Generally, the primary study selection task is formulated as a supervised learning problem~\citep{Adeva2014,Almeida2016}. Some authors adopt binary classification to discern between relevant or non-relevant papers, while others apply active learning~\citep{Wallace2012,Yu2019}, which includes a human oracle to confirm the relevance of a sample of papers. 
The more recent advances in the field are related to the use of LLMs to elucidate whether a candidate paper is within scope~\citep{Castillo-Segura2023} and should be selected~\citep{Thode2025}.

However, current approaches have some limitations. Active learning techniques require manual paper review, a process that can lead to inconsistencies if a large number of papers are presented to the user for confirmation. In this case, the benefits of automation are greatly reduced. In practice, some studies replace the human oracle with an automatic oracle based on available data labels. As for supervised techniques, most authors use algorithms such as support vector machines (SVM) and neural networks~\citep{DeLaTorre2023}. The reason seems to be their ability to cope with the high dimensionality of the data, since the features used for training are extracted by text mining. Consequently, black-box models are built that hide any patterns that might help the researcher understand why a paper is selected or discarded. A third limitation comes from the information extracted from the papers to feed the training process. Most proposals use frequent terms extracted from the title, abstract and keywords. Although textual information is usually considered for manual selection, other aspects of the paper could be decisive in assessing not only its relevance to the research topic, but also its quality. Finally, LLM-based studies remain exploratory in nature, with a heavy reliance on volatile and expensive technology. While LLMs can handle textual information more intuitively, they still lack transparency and exhibit nondeterministic behaviour. Ultimately, they may require more human supervision to confirm the relevance of automatically selected articles.

In this paper, we address these limitations and propose a new approach for automatic paper selection based on evolutionary ML~\citep{Telikani2021}: \ourmodel (Interpretable Rule-based Evolutionary Classification for primary study Selection). A grammar-guided genetic programming (G3P) algorithm first generates class association rules (CARs) guided by a fitness function specially designed to cope with the data imbalance. Next, the rules are selected to build a rule-based classifier, which is recognised as a more interpretable model. The mined rules describe conditions ---in form of logic predicates--- that papers satisfy or not to be classified as relevant. Apart from textual information, we incorporate bibliometric information such as number of citations, number of authors or type of publication, which can provide additional useful information to properly classify papers. The context-free grammar provides high flexibility to decide the structure of the rules, as well as the logic operators and sources of information that serve to define them.

We perform several experiments to evaluate the contribution of the novel components of our proposal. First, a parameter study analyses the influence of standard evolutionary parameters as well as the associative rule selection strategy for classifier construction. Next, we study the influence of bibliometric information as part of the rule generation process. We also analyse the interpretability of the rules by focusing on the combination of textual and bibliometric information. In addition, we compare our approach with FAST2, a recent proposal based on active learning~\citep{Yu2019}. Our results show that \ourmodel performs better on classification metrics and improves the interpretability of the output as a result of the classifier rules.

The remainder of the paper is organised as follows. Section~\ref{sec:background} introduces relevant concepts related to automatic paper selection, and the basics of associative classification and GP3. Section~\ref{sec:method} details the design of \ourmodel. The experimental framework to evaluate \ourmodel is defined in Section~\ref{sec:framework}. Section~\ref{sec:results} presents the results for each research question. Conclusions and future work are outlined in Section~\ref{sec:conclusions}.

\section{Background and related work}
\label{sec:background}

This section presents the algorithmic foundations of associative classification (Section~\ref{subsec:associative-classification}) and G3P (Section~\ref{subsec:g3p}), as well as related work on automatic paper selection with ML (Section~\ref{subsec:relWork}).

\subsection{Associative classification}
\label{subsec:associative-classification}

Association rule mining (ARM)~\citep{Agrawal1993, Zhang2002} is an unsupervised learning paradigm that allows discovering interesting patterns in the form of rules ($A \to C$). This type of rule expresses that when the conditions that compose the antecedent ($A$) are given, the consequent ($C$) also happens. $A$ and $C$ are disjoint itemsets, \ie $A\;{\displaystyle \cap }\;C = \emptyset$, where each item is usually a logic expression of the form: $<attributte><logic\_operator><value>$.

Thus, ARM techniques can extract rules to describe patterns in a dataset. To evaluate the quality of the rules, two common metrics are support and confidence. Support counts the proportion of transactions ($T$) in the entire dataset ($D$) that satisfy parts $A$ and $C$ of the rule (see equation~\ref{eq:support}). Confidence measures the reliability of the rule as it divides the number of transactions satisfying $A$ and $C$ by the number of transactions where $A$ is satisfied (see Equation~\ref{eq:confidence}).

\begin{equation}
support(A \to C) = \frac{|\{A \;{\displaystyle \cap}\;C \subseteq T, T \in D\}|}{|D|}
\label{eq:support}
\end{equation}

\begin{equation}
confidence(A \to C) = \frac{support(A \to C)}{support(A)}
\label{eq:confidence}
\end{equation}

Rules extracted using ARM techniques can be used to build accurate classifiers capable of solving prediction problems~\citep{Antonie2004}. A rule-based classifier defines a set of rules that, when applied in a specific order to an instance of the dataset, assigns it to a category (class). CARs are specific rules that require the consequent to contain only the class attribute and its value, \ie $<class>=<value>$. The standard procedure for constructing associative classifiers is to mine the dataset to find a set of CARs with minimal support and confidence~\citep{Lui1998}. Second, a classifier is constructed by inspecting the rules and choosing the appropriate ones. As an optional third phase, pruning can be applied to reduce the number of rules.

To build an accurate classifier, it is important to correctly choose a subset of the extracted rules and rank them. Different strategies have been proposed in the literature for this purpose~\citep{Thabtah2007}. The most common classification algorithm is CBA (Classification Based on Associations)~\citep{Lui1998}. CBA sorts the generated rules based on their coverage in the dataset, \ie how many instances the rule can be applied to. It then iterates rule by rule checking if the antecedent is satisfied. If so, the rule is marked to be part of the classifier. The process stops once all instances in the dataset are covered by at least one rule. The order of the rules in the final classifier is usually determined by support or confidence.

In the following years, alternatives to CBA were proposed. CMAR (Classification based on Multiple Association Rules)~\citep{Li2001} relies on FP-trees to improve efficiency, pruning rules based on confidence, correlation and coverage. Another proposal is CPAR (Classification based on Predictive Association Rules)~\citep{Yin2003}, which uses a statistical measure, called Laplace expected error estimate, to predict rule accuracy. Finally, SCBA (Scoring based Association Rules)~\citep{Chen2012} calculates rule scores based on a weighted confidence of positive instances and negative instances. 

\subsection{Grammar-guided genetic programming}
\label{subsec:g3p}

Evolutionary computation (EC) encompasses a family of bio-inspired algorithms based on the Darwinian theory of natural evolution~\citep{Eiben2015}. In EC, individuals, which are possible solutions to an optimisation problem, are encoded and represented by their genotype. After the initial generation of a population of individuals, an iterative procedure is initiated. In each generation, individuals undergo selection, recombination, mutation and replacement operators with the ultimate goal of finding the best solution. A key aspect in the design of an evolutionary algorithm is the definition of the fitness function. This problem-specific function evaluates the quality of the individual and is applied in various parts of the algorithm to compare individuals and determine which should survive.

Grammar-guided genetic programming (G3P) is a particular EC paradigm in which the individual genotype represents a derivation tree of a context-free grammar (CFG)~\citep{McRay2010}. The CFG defines the shape and constrains the size of the tree, thus allowing better control of the possible solutions to a problem. The CFG is defined by the following tuple $\{S, \sum_N , \sum_T , P\}$, where $S$ is the root symbol, $\sum_N$ represents the alphabet of non-terminal symbols, $ \sum_T$ is the alphabet of terminal symbols, and $P$ is the set of production rules. The production rules describe how non-terminal symbols can be derived into other non-terminal or terminal symbols. It is important to note that $\sum_N \;{\displaystyle \cap}\; \sum_T = \emptyset$. The derivation tree, which always starts with the root symbol $S$, is the result of applying randomly chosen production rules, expanding its nodes until the leaves contain only terminal symbols. The leaves of the tree, read in preorder, represent the phenotype of the individual, \ie the real-world representation of the genotype. The main advantage of using a CFG is that the algorithm always provides valid solutions to the problem, since they stick to the symbols and rules of the grammar.

G3P can be applied to mine association rules~\citep{Luna2012} and CARs~\citep{Barbudo2021}, using the CFG to declare the logic operators and the structure of the rules. In this case, each individual symbolises a rule, whose fitness is usually evaluated in terms of support and confidence. As a result, a set of the best rules describing the input dataset is returned.

\subsection{Automatic selection of primary studies}
\label{subsec:relWork}

Finding the most relevant publications to a specific research topic requires a thorough search procedure and extensive reading. Within an SLR, the selection of primary studies is probably the most costly task both in time and resources~\citep{OMara-Eves2015}. It also requires research experience and a systematic method to avoid bias. Some tools can assist researchers in this process~\citep{Marshall2013}, offering functionalities to collect and visualise the selected papers as graphs.

Focusing on AI techniques to support the automation of paper selection, we discuss below the most relevant ML approaches~\citep{DeLaTorre2023}. It should be noted that the nature of the problem requires the combination of ML with text mining to extract textual features typically used for the learning phase. TD-IDF (Term Frequency-Inverse Document Frequency) and bag-of-words (BoW) are common approaches in text mining for this task. The pioneering ML work is based on voting perceptrons and text mining~\citep{Cohen2006}. However, the vast majority of the proposals in the literature adopt SVM~\citep{Kim2014,Bannach-Brown2019,Thomas2021}. Other supervised techniques have been explored, such as Naive Bayes \citep{Matwin2010}, logistic regression \citep{Bannach-Brown2019} or decision trees \citep{Langlois2018}. 
A recent study compares the performance of SVM and random forest with human-driven paper selection in SLR updates~\citep{Costalonga2025}. The authors conclude that ML models help reduce the paper selection effort, but the obtained F-score (0.33) was too low. However, it should be noted that they address a different problem than ours, since an SLR update involves two steps of search, screening, and selection.

Focusing on active learning, we should mention two relevant methods: Abstractkr~\citep{Wallace2012} and FAST2~\citep{Yu2019}. Abstrackr is a web tool that displays a sample of papers to the researcher, who must label them as relevant or not based on some abstract terms highlighted by the tool. Internally, Abstrackr builds a predictive model using SVM to automatically determine the relevance of the remaining papers. FAST2 uses text mining to explore relevant terms in the dataset. Its SVM classifier is updated after a new set of 10 papers is labelled by a human or automatic oracle. FAST2 is designed to achieve 95\% recall with the smallest number of labelled papers. The authors provide the code in a public repository.

The most recent advances have explored the use of LLMs for paper selection. In 2023, \cite{Castillo-Segura2023} analysed six LLMs, asking them to decide whether the abstract of a candidate paper fit the scope of an SLR topic. The prompts included some keywords to contextualize the SLR topic, requesting a yes/no response and two reasons to justify the classification. The accuracy was measured using 596 abstracts manually classified by the authors. The recall results (between 20\% and 55\%) appear to be far from those obtained with supervised techniques. Another study compares five LLMs, also considering peer collaboration in paper selection~\citep{Thode2025}. In this case, the LLMs were provided with the inclusion and exclusion criteria of two SLRs. Based on these criteria, the LLMs were able to decide whether candidate papers should be selected after analysing their title and abstract. Their experimental results show that ensuring both recall and precision is a difficult tasks. While some LLMs reached over 95\% recall, precision was generally below 50\% due to a high number of false positives. The recall values increase when combining two LLMs, as the approach is to accept the paper as relevant if at least one LLM selects it. In contrast, precision does not experience such an improvement, and values remain low.

\section{Proposed approach} 
\label{sec:method}

\begin{figure}
\centering
\includegraphics[width=\textwidth]{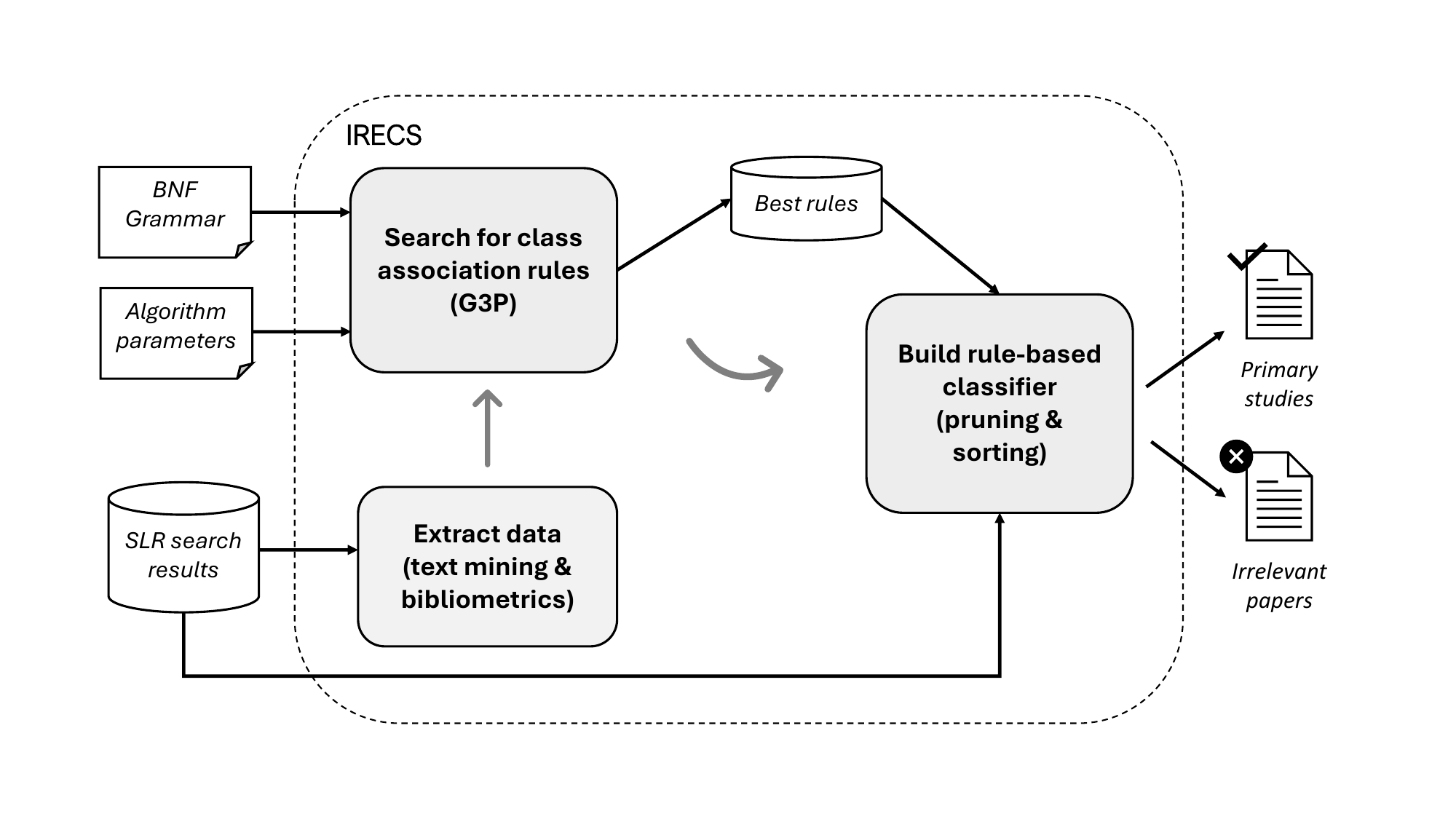}
\caption{Overview of \ourmodel components, inputs and outputs.}
\label{fig:g3p4slr}
\end{figure}

Figure~\ref{fig:g3p4slr} shows an overview of \ourmodel, our proposed approach for automatic paper selection. It consists of three main components: a data extraction module that uses text mining to create the dataset for training, a G3P algorithm to mine CARs, and the procedure to build the rule-based classifier. The inputs are: the CFG, specified in BNF format; the parameter settings of the G3P algorithm; and a corpus of papers resulting from a literature search. The G3P configuration file allows customising parameters such as the maximum number of generations, population size, the probabilities of the genetic operators, and the strategy for choosing the rules for building the classifier. Note that the algorithm has been implemented to be highly configurable, and default values are provided. However, it seems desirable to adjust the parameters depending on the SLR corpus, as explained in our parameter study (see Section~\ref{subsec:RQ1-results}). As for the corpus, our proposal requires title and abstract to extract textual information. In addition, bibliometric information on each paper would be highly recommended to enrich the CFG. We currently include the following metrics: year of publication, number of citations, number of papers and type of publication (journal or conference). We automatically retrieve them from Scopus using the paper's DOI.

\begin{algorithm}[ht]
\caption{\ourmodel algorithm for rule generation}\label{alg:cap}
\setcounter{AlgoLine}{0}
\SetKwInOut{Input}{Input}
\SetKwInOut{Output}{Output}
\Input{dataset, cfg}
\Output{selectedRules}
$vocabulary \gets extractVocabulary(dataset)$\\
$rules \gets findRules(dataset, cfg, vocabulary)$\\
$sortedRules \gets sort(rules)$\\
$selectedRules \gets \emptyset $\\
$uncoveredInstances \gets dataset$\\
\ForEach {rule in sortedRules}{
    \ForEach {instance in uncoveredInstances} {
    \If {covers(instance, rule)}{
    $uncoveredInstances \gets uncoveredInstances - instance$ \\
    $selectedRules \gets selectedRules \cup rule$\\
    }
    \If{empty(uncoveredInstances)}{
        break\\
    }
    }
}
\end{algorithm}

Algorithm~\ref{alg:cap} summarises the internal steps of the process, from data extraction to rule selection. The first step involves vocabulary extraction from the SLR corpus by text mining. We apply TD-IDF and Porter stemmer to extract the most relevant words from the title and abstract of each paper. The vocabulary will be used by some grammar operators to express conditions on the papers. For instance, we have implemented a logic operator called ``contains'' to determine whether the title of a paper contains a specific word. It would allow us to generate rules with the following structure: $IF\ <title>\ <contains>\ <word>\ THEN\ <isCandidate><=><``True''>,\ word \in vocabulary$. This type of rule is very easy to read and interpret, and the user could even specify his/her own vocabulary of interest. This idea provides much more flexibility and interpretability than the large feature set (one per frequent word) used by SVM-based proposals. 

Once the vocabulary is ready, \ourmodel starts the evolutionary rule generation process for the configured number of generations. This process is detailed in Sections~\ref{subsec:grammar-encoding} to~\ref{subsec:archive}. The output is a set of CARs that can be used to classify papers, based on the patterns observed in the training partition of the corpus. To build the classifier, rules are sorted by fitness and its coverage is determined. Next, the strategy to select the rules is applied. For this step, we will study the performance of the strategies explained in Section~\ref{subsec:associative-classification} as part of our experimentation. To evaluate the accuracy of the resulting classifier, the test partition of the corpus is required.

\subsection{Grammar and encoding}
\label{subsec:grammar-encoding}

Following the precepts of G3P, individuals are encoded as derivation trees that conform to a CFG. The grammar participates in two moments of the evolutionary search: initialisation of the population, to create individuals from scratch; and mutation, when some parts of the individual genotype are transformed. A completely new individual (rule) is constructed by a random derivation process that starts with the root symbol and ends when all non-terminal symbols have been derived into terminal ones. The CFG defined for the primary study selection problem, shown in Figure~\ref{fig:grammar}, is intended to produce CARs. The root symbol ($<rule>$) always derives into an antecedent ($<antc>$) and a consequent ($<consq>$). The consequent always derives into a single categorical comparator, which can be either $equals$ or $notEquals$, followed by the operator \texttt{isCandidate} and the terminal symbol \texttt{candidateStudyValue} (a boolean value).

\begin{figure}[!t]
\begin{center}
\begin{lstlisting}[xleftmargin=0.0\textwidth, xrightmargin=0.0\textwidth]
(*@$S$@*) = <rule>

(*@$\sum_N$@*) = {(*@<rule>,@*) (*@<antc>,@*) (*@<consq>,@*) (*@<cmp>,@*) (*@<numCmp>,@*) (*@<textCmp>,@*) (*@<catCmptor>,@*) (*@<numCmptor>,@*) (*@<textCmptor>@*)}

(*@$\sum_T$@*) = {(*@\textit{not}@*), (*@\textit{and}@*), (*@\textit{or}@*),  (*@\textit{equals}@*) (*@\textit{notEquals}@*) (*@\textit{>}@*), (*@\textit{<}@*), (*@\textit{>=}@*), (*@\textit{<=}@*), (*@\textit{containsAll}@*), (*@\textit{containsAny}@*),(*@\textit{nCites}@*), (*@\textit{nAuthors}@*), (*@\textit{year}@*), (*@\textit{title}@*), (*@\textit{abstract}@*), (*@\textit{titleAbstract}@*), (*@\textit{paperType}@*), (*@\textit{isCandidate}@*), (*@\textit{candidateStudyValue}@*), (*@\textit{numValue}@*), (*@\textit{nCitesValue}@*), (*@\textit{nAuthorsValue}@*), (*@\textit{yearValue}@*), (*@\textit{titleValue}@*), (*@\textit{abstractValue}@*), (*@\textit{titleAbstractValue}@*), (*@\textit{paperTypeValue}@*)}

(*@$P$@*) = {

 <rule>       ::= <antc> <consq>
 <antc>       ::= <cmp> | (*@\textit{not}@*) <cmp> | (*@\textit{and}@*) <cmp> <antc>

 <consq>      ::= <catCmptor> (*@\textit{isCandidate}@*) (*@\textit{candidateStudyValue}@*)

 <cmp>        ::= (*@\textit{and}@*) <numCmp> <textCmp> | <textCmp> 

 <numCmp>     ::= <numCmptor> (*@\textit{nCites}@*) (*@\textit{nCitesValue}@*)
     | <numCmptor> (*@\textit{nAuthors}@*) (*@\textit{nAuthorsValue}@*)
     | <numCmptor> (*@\textit{year}@*) (*@\textit{yearValue}@*)

 <textCmp>    ::= <textCmptor> (*@\textit{title}@*) (*@\textit{titleValue}@*)
     | <textCmptor> (*@\textit{abstract}@*) (*@\textit{abstractValue}@*)
     | <textCmptor> (*@\textit{titleAbstract}@*) (*@\textit{titleAbstractValue}@*)
     | <textCmptor> (*@\textit{paperType}@*) (*@\textit{paperTypeValue}@*)

 <catCmptor>  ::= (*@\textit{equals}@*) (*@\textit{notEquals}@*)
 
 <numCmptor>  ::= (*@\textit{>}@*) | (*@\textit{<}@*) | (*@\textit{>=}@*) | (*@\textit{<=}@*)

 <textCmptor> ::= (*@\textit{containsAll}@*) | (*@\textit{containsAny}@*)      
}
\end{lstlisting}
\end{center}
\caption{The context-free grammar defined to generate class associative rules to classify scientific papers into relevant (candidate) or not.}
\label{fig:grammar}
\end{figure}

The derivation process of the antecedent leads to one or more logical expressions ($<cmp>$), concatenated by means of the $and$ operator. The negation of an expression by $not$ is also possible. Each condition can use a numeric ($>,<,\ge, \le$) or textual ($containsAll$ or $containsAny$) comparator. Numerical comparisons are intended to reflect patterns based on bibliometric information. Number of citations ($nCites$), number of authors ($nAuthor$) and publication year ($year$) are the currently implemented options. On the other hand, title ($title$), abstract ($abstract$), the combination of both ($titleAbstract$) and publication type ($paperType$) serve to construct categorical comparisons based on textual information obtained from the papers. For terminal symbols ending in ``value'', the grammar configuration specifies the range of possible values. For instance, we can limit the range for the number of authors using the following XML syntax:

\begin{equation}
\begin{split}
<terminal name="nAuthorsValue" code="nAuthorsValue" \\
type="int" minValue="1" maxValue="20" />.
\end{split}
\end{equation}

\begin{figure}[ht]
    \centering
    \includegraphics[width=0.95\linewidth]{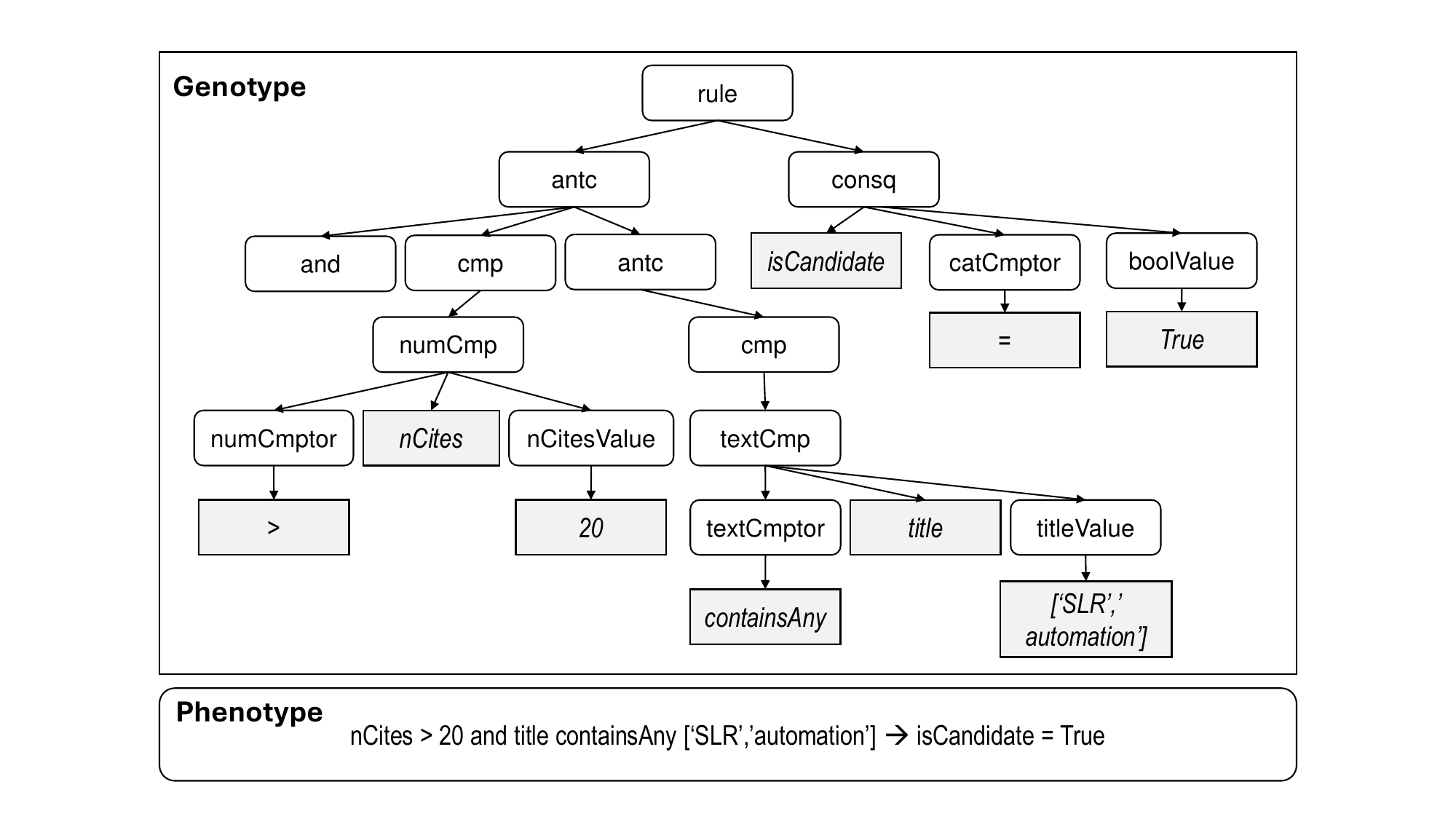}
    \caption{An example individual represented by its genotype (derivation tree) and phenotype (class association rule).}
    \label{fig:genotypePhenotypeRule}
\end{figure}

This XML fragment fully defines how the terminal symbol is to be substituted for a specific value, as it includes a type declaration (integer) and the range of values among which the algorithm will make random choices. The code attribute links the symbol to the name of the function to be executed to obtain that value. In the case of categorical values, each has its own logic (code function) to determine the value to be substituted for. The vocabulary is needed to compare title and abstract, while the article type includes specific categories such as ``journal'' or ``conference''. Figure~\ref{fig:genotypePhenotypeRule} shows an example of a derivation tree (genotype) and the rule it represents (phenotype). In the derivation tree, terminal symbols are shaded. In this case, the antecedent concatenates two expressions, a numeric and a textual one, to suggest that a paper with more than 20 citations and whose title contains the terms ``SLR'' or ``automation'' should be considered as a relevant paper.

\subsection{Fitness function}
\label{subsec:fitness-function}

The fitness function is the core element of the G3P algorithm, as it is used in several steps to compare solutions. Existing G3P proposals to rule mining, such as those mentioned in Section~\ref{subsec:g3p}, use the $support$ metric as the fitness function. However, our initial experiments using this metric did not yield competitive results. Rules with high support ensure that they are likely to cover many instances in the dataset, but this does not necessarily mean they are good for classification. SLR datasets can be highly imbalanced, \ie most of papers retrieved by an automatic search are not ultimately selected as primary studies. Imbalanced datasets require a more problem-oriented fitness function to cope with the class distribution. Otherwise, the search process will be unable to find good rules to describe the distinguishing characteristics of positive instances. 

We have designed a specific fitness function that considers whether the consequent of the rule refers to positive or negative instances. The intuition is that we need to promote rules that are good at describing only positive instances or that are only satisfied by negative instances. Very generic rules that cover both positive and negative instances are not desirable, as they will not help the classifier to discern between both classes. The proposed fitness function is formulated in Equation~\ref{eq:fitness}. If the rule consequent includes the condition $isCandidate=``True"$ ($R^+$), we count how many instances match the rule among instances labelled as positive. The second term penalises the fitness value if negative instances are also covered by the rule antecedent, as it will result in misclassification. The opposite idea applies to rules ending in $isCandidate=``False"$ ($R^-$).

\begin{multline}
fitness(R^+) = \frac{|InstancesMatchRule|}{|PosInstances|} - \frac{|NegInstancesMatchAntecedent|}{|NegInstances|}\\
fitness(R^-) = \frac{|InstancesMatchRule|}{|NegInstances|} - \frac{|PosInstancesMatchAntecedent|}{|PosInstances|}
\label{eq:fitness}
\end{multline}

\subsection{Genetic operators}
\label{subsec:genetic-operators}

The selection operators chooses individuals that will act as parents for the generation of offspring solutions. In our algorithm, binary tournament selection is applied. For each pair of parents, a crossover operator is run to recombine their genotypes with a given probability. We follow the standard procedure in G3P, which consists of randomly choosing a common non-terminal symbol. Then, the operator swaps the derivation subtrees starting in the selected non-terminal symbol.

As for mutation, it applies a transformation to a single individual also with a certain probability. A random node is chosen from the tree. If it corresponds to a non-terminal symbol, a new derivation process is activated to reconstruct that branch. For terminal symbols, it is sufficient to replace the value by an equivalent one according to the type of operator. Once individuals are recombined and possibly mutated, a new population has to be created to continue the evolution. We implement three alternatives to experimentally study its performance in the search process. A first method (NPOP) simply replaces the current population with a random sample of all solutions (current population and descendants) to enhance diversity. The second method (ELIT) applies an elitism strategy that keeps the best individuals among the current population and the offspring. Lastly, the pure generational approach (PGEN) promotes the survival of the new individuals created in the previous generation.

\subsection{Archive of best rules}
\label{subsec:archive}
Standard single-objective G3P algorithms evolve a population of solutions, but only return the best individual at the end of the search process. When applied to CAR generation, a set of good candidate rules is needed for classifier construction. For this purpose, a secondary population is defined that stores the best rules found so far~\citep{Barbudo2021}. The purpose of this archive of best rules is twofold. The evolutionary algorithm does not lose the best individuals if a strong diversity-oriented replacement strategy is applied. Moreover, it allows to collect a sufficient number of rules to build the classifier without interfering with a wide exploration of the search space. We adopt this idea in \ourmodel. Initially, the archive of best rules is empty. After each generation, rules exceeding a fitness threshold are included in the archive. Rules stored in the archive at the end of the search are returned and undergo rule sorting and pruning for classifier construction.

\section{Experimental framework}
\label{sec:framework}

In this section, we formulate the research questions (RQ) to be experimentally validated. We also detail the experimental methodology (Section~\ref{subsec:methodology}), the datasets obtained from SLR studies (Section~\ref{subsec:datasets}), and the evaluation metrics to assess the classifier accuracy and interpretability (Section~\ref{subsec:metrics}).

\subsection{Research questions}
\label{subsec:rqs}

Four RQs are proposed to validate different aspects of our proposal:

\begin{itemize}
    \item \textit{RQ1: Which parameter configuration offers the best performance for automatic classification of relevant studies?} \ourmodel includes several parameters that influence the search for CARs and how the classifier is built. More specifically, we want to analyse the impact of the replacement strategy, the parameters of the evolutionary search, and the classification and pruning strategy. Our parameter study will allow us to discover whether these parameters and components require a different configuration depending on the dataset, and to recommend a general-purpose configuration with good performance.

    \item \textit{RQ2: Do bibliometric operators contribute to improved classification performance?} One of the distinguished features of our proposal with respect to the state of the art is the use of data features beyond the keywords extracted from title and abstract. In this RQ, we seek to prove that the consideration of bibliometric information is useful to improve classification results. To this end, we will compare the performance of \ourmodel with and without activating specific grammar operators that rely on bibliometric information.

    \item \textit{RQ3: What are the characteristics of the rules selected to build rule-based classifiers for automatic paper selection?} Another relevant aspect of \ourmodel is the generation of a rule-based classifier, which is known to be more interpretable than existing approaches based on SVM or neural networks. In this RQ, we address the study of rule interpretability with the goal of helping researchers understand the information used to identify relevant studies.
    
    \item \textit{RQ4: How does \ourmodel perform compared to a state-of-the-art black box classifier?} To contextualise the advantages of our approach, we study its performance in comparison with FAST2, 
    an algorithm for automatic paper selection based on active learning~\citep{Yu2019}. In this RQ, we analyse the strengths and limitations of both methods based on their results on various datasets and classification metrics.
\end{itemize}

\subsection{Methodology}
\label{subsec:methodology}

Our approach is based on EC, whose stochastic nature implies that several runs must be performed. In our experiments, we run the algorithm with different random seeds. More specifically, we perform 10 runs for all the experiments. Another general methodological decision is the application of k-fold cross validation for rule extraction and classifier construction. We adopt $k=5$ and a stratification approach due to the class imbalance.

Next, we detail specific decisions for the experiments designed to validate each RQ. For RQ1, we need to select the components of \ourmodel to be analysed and the specific values for its parameters. The assessment of the configurable elements is done in three steps. First, we analyse the influence of the classification construction procedure based on rule pruning and sorting. We consider the four strategies explained in Section~\ref{subsec:associative-classification}: CBA, CMAR, CPA and SCBA. After selecting the strategy, we focus on the evolutionary search, studying the influence of the three replacement strategies implemented: NPOP, that randomly creates a new population in each generation; ELIT, that promotes the survival of the best individuals according to their fitness; and PGEN, that replaces the current population by the new descendants. Finally, Table~\ref{tab:parameters} shows the values of the parameters controlling the population size ($popSize$), the number of generations ($maxGen$), and the probabilities of the genetic operators ($crossProb$ and $mutProb$). To evaluate the pruning and replacement strategies, we set the G3P parameters to the following default values: $maxGen=100$, $popSize=100$, $crossProb=0.9$ and $mutProb=0.1$. The maximum number of derivations, which control the size of the rules, is set to 15. The fitness threshold is specific for each dataset.\footnote{The specific values can be consulted in the replication package.} We set all these values after preliminary experiments. In total, the number of configurations under comparison is 16.

\begin{table}[ht]
\centering
\begin{tabular}{lllll}
\hline
maxGen & popSize & crossProb & mutProb\\
\hline
100     & 100   & 0.9   & 0.1 \\ 
50      & 200   & 0.9   & 0.1 \\ 
200     & 50    & 0.9   & 0.1 \\
100     & 100   & 0.7   & 0.05\\
50      & 200   & 0.7   & 0.05\\ 
200     & 50    & 0.7   & 0.05\\
\hline
\end{tabular}
\caption{Evaluated parameter configurations of the G3P algorithm for mining CARs.}
\label{tab:parameters}
\end{table}

For RQ2, RQ3 and RQ4, we use a single configuration of our algorithm, which performs adequately for all datasets. We consider such a configuration as the default for a general purpose. In RQ2, we compare two versions of \ourmodel: one using the full grammar definition (see Figure~\ref{fig:grammar}) and one without the production rules and symbols related to bibliometric information. In this way, we can evaluate the contribution of this type of information in rule mining and classification performance. In this experiment, the number of datasets must be reduced, since not all of them provide this type of information (see details in Section~\ref{subsec:datasets}).

Regarding RQ4, the choice of FAST2 over more recent approaches is justified by two main reasons. First, the FAST2 code is publicly available, as are the datasets used for experimentation. Second, FAST2 and \ourmodel can be configured to use the same features for training, which defines a fair comparison scenario. Despite this, the fact that FAST2 follows an active learning approach imply certain methodological decisions. 
FAST2 requires an oracle to label some papers at regular intervals. In this regard, we enable its simulation mode in which an automatic oracle provides the true label included in the dataset, thus avoiding human labelling. In addition, FAST2 is designed to stop the process once 95\% recall is reached, and does not perform cross-validation. In this case, we adapted the execution procedure to adopt the same training/testing phases required by \ourmodel. In this way, the internal classifier built by FAST2 can be evaluated under the same conditions as ours, extracting classification performance metrics on the test partition. We apply the Wilcoxon test ($alpha=0.05$) to each performance metric to assess the difference between the two algorithms. For this test, the null hypothesis establishes that both algorithms perform equally.

\subsection{Datasets}
\label{subsec:datasets}

We use five datasets to conduct our experiments, whose characteristics are shown in Table~\ref{tab:datasets}. Each dataset represents a collection of research papers returned in the search process of an authentic SLR study. Each instance contains the textual and bibliometric information (if possible) of a paper, together with the human-annotated label specifying whether it was ultimately included in the set of primary studies. The datasets come from three different sources. Hall, Wahono and Kitchenham datasets are obtained from the FAST2 replication package~\citep{Yu2019}. The Muthu dataset is available on Github\footnote{https://github.com/asreview/synergy-dataset} as part of the collection provided by the ASReview tool~\citep{VanDeSchoot2021}. Finally, the AISE dataset is a new dataset that we created from a literature search of a survey we conducted in the past~\citep{Ramirez2019}. Four datasets refer to software engineering topics, while the Muthu dataset focuses on medicine. With the exception of AISE, all datasets are highly imbalanced. The percentage of candidate papers selected as primary studies is less than 3\%.

\begin{table}
\centering
\begin{tabular}{lllllll}
\hline
Dataset & Research & Biblio. & Pos. & Neg. & Total & Class\\
& field & info. & instances & instances & instances & distr.\\
\hline
AISE & Soft. Eng. & Yes & 5281 & 19521 & 24802 & 21.29\% \\
Muthu & Medicine & Yes & 13 & 442 & 455 & 2.85\%\\
Hall & Soft. Eng. & Yes & 106 & 8805 & 8911 & 1.1\% \\
Kitchenham & Soft. Eng. & No & 4 & 1659 & 1704 & 2.64\%\\
Wahono & Soft. Eng. & No & 62 & 6940 & 7002 & 0.89\%\\
\hline
\end{tabular}
\caption{Characteristics of the SLR datasets.}
\label{tab:datasets}
\end{table}

Originally, the publicly available datasets did not contain any bibliometric information. We implemented and executed API queries to Scopus to obtain the bibliometric information for each instance. We retrieved the number of authors, type of publication (journal or conference) and number of citations. The years of publication were already included in the original datasets, although they had never been used in previous works. Our automated process was not able to find the information for some cases, so a manual review was performed to complete the missing data. In the case of two datasets, the amount of missing information was extremely large, so we kept them as original due to the high cost of completing them manually. These datasets are not taken into account for RQ2. Similarly, since FAST2 only supports textual features, we used the version of the datasets without bibliometric information for RQ4. In fact, our algorithm disables grammatical operators and symbols related to bibliometric information if the corresponding data features are not present in the given dataset.

For textual features, we not only need to extract the relevant keywords from the title and abstract of each paper ---as other proposals do--- but also identify a global vocabulary. This set of terms is needed for the $containsAll$ and $containsAny$ grammar operators. For the textual information of each paper, we apply the same procedure as in FAST2, which applies TF-IDF. This technique assists in identifying term's relevance in a document by taking into account how often it occurs in the text as well as how uncommonly it occurs throughout the collection of documents. Higher TF-IDF scoring terms are thought to be more significant to the text. For vocabulary definition, we also apply text mining, but considering positive and negative instances independently. Therefore, we first obtain two vocabularies: $Vocabulary_{POS}$, which contains frequent terms in the title and abstract of the positive instances (selected papers); and $Vocabulary_{NEG}$, which applies the same idea to the negative instances (non-selected papers). The final vocabulary is obtained as $Vocabulary_{RELEVANT} = Vocabulary_{POS} - Vocabulary_{NEG}$. We hypothesise that relevant words appearing in positive but not in negative instances would help the classifier to discern between the two classes. In fact, in our preliminary experiments it gave better results than defining a global vocabulary from all papers indiscriminately. It should be noted that users could also define their own vocabulary manually, which would provide even more guidance to the search. Our automatic approach to some extent simulates what a human would do, since it reinforces the occurrence of keywords that are unique to the relevant papers and discards those that might lead to confusion because they are too general.

\subsection{Evaluation metrics}
\label{subsec:metrics}

The performance of algorithms and configurations in RQ1, RQ2 and RQ4 is measured by standard classification metrics: recall (Equation~\ref{eq:rec}), precision (Equation~\ref{eq:pre}), specificity (Equation~\ref{eq:spec}), and accuracy (Equation~\ref{eq:acc}), where TP are true positives, TN are true negatives, FP are false positives, and FN are false negatives. Due to the highly imbalanced nature of most of the datasets, we also compute the balanced accuracy (Equation~\ref{eq:bal-acc}). For \ourmodel, we report the average results of the runs with different random seeds and data folds.

\begin{equation}
recall = \frac{\text{TP}}{\text{TP} + \text{FN}}
\label{eq:rec}
\end{equation}

\begin{equation}
precision = \frac{\text{TP}}{\text{TP} + \text{FP}}
\label{eq:pre}
\end{equation}

\begin{equation}
specificity = \frac{\text{TN}}{\text{TN} + \text{FP}}
\label{eq:spec}
\end{equation}

\begin{equation}
accuracy = \frac{\text{TP+TN}}{\text{TP+TN+FP+FN}}
\label{eq:acc}
\end{equation}

\begin{equation}
bal-accuracy = \frac{1}{2}(\frac{\text{TP}}{\text{TP+FN}} + \frac{\text{TN}}{\text{TN+FP}})
\label{eq:bal-acc}
\end{equation}

For RQ3, we report additional measures related to interpretability: rule length, \ie number of conditions in the antecedent; number of bibliometric operators appearing in the rules; and the number of rules comprising the rule-based classifier.


\section{Results and discussion}
\label{sec:results}

This section presents and analyses the experimental results that serve to answer each RQ.

\subsection{RQ1: Parameter study}
\label{subsec:RQ1-results}

\begin{figure}[ht]
    \centering
    \subfloat[CBA\label{subfig:cba}]{%
      \includegraphics[height=4cm]{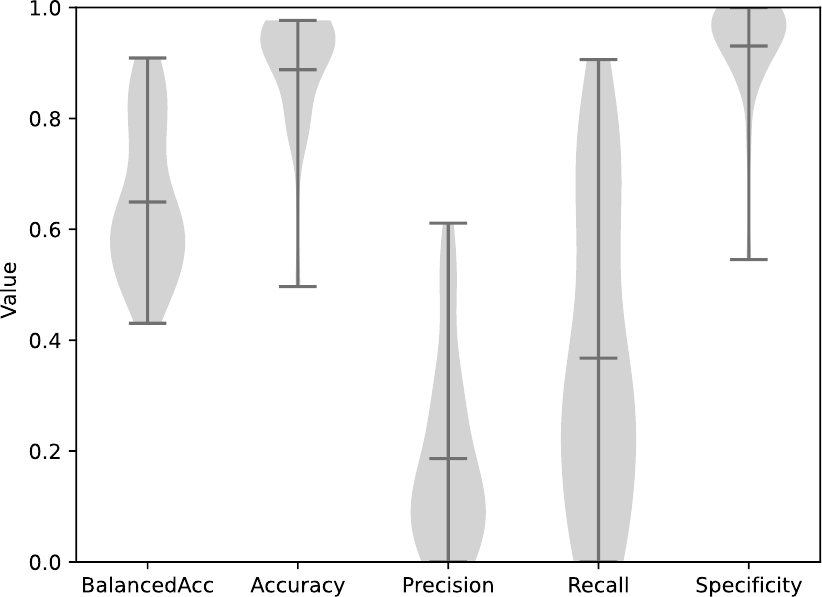}
    }
    \subfloat[SCBA\label{subfig:scba}]{%
      \includegraphics[height=4cm]{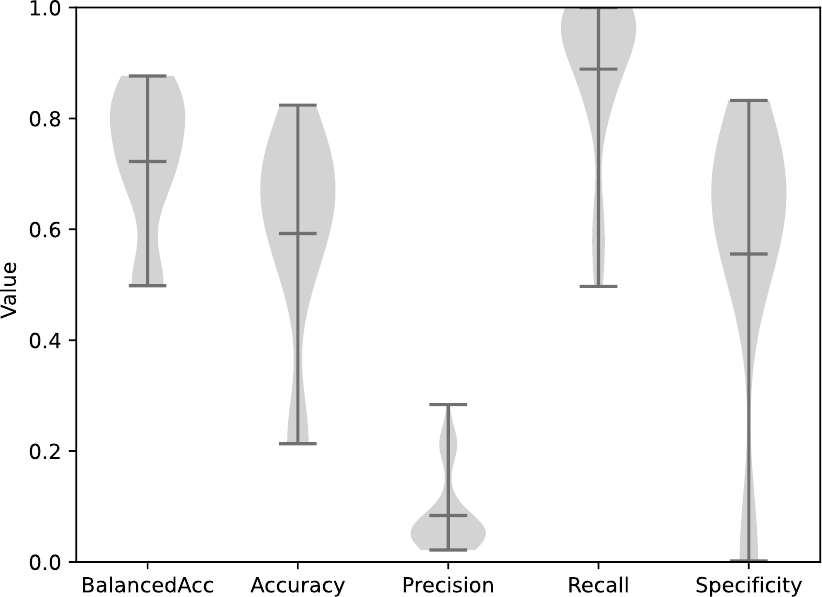}
    }
    \\
    \subfloat[CPAR\label{subfig:cpar}]{%
      \includegraphics[height=4cm]{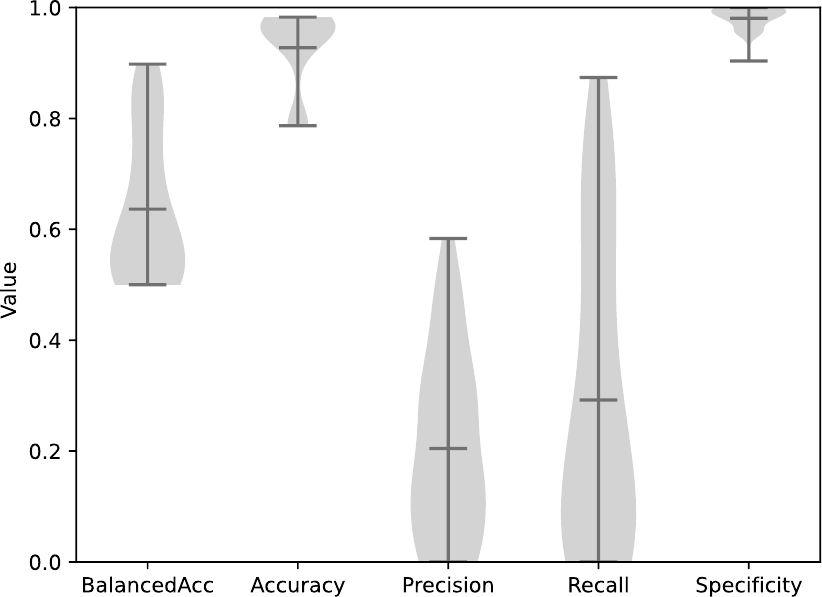}
    }
    \subfloat[CMAR\label{subfig:cmar}]{%
      \includegraphics[height=4cm]{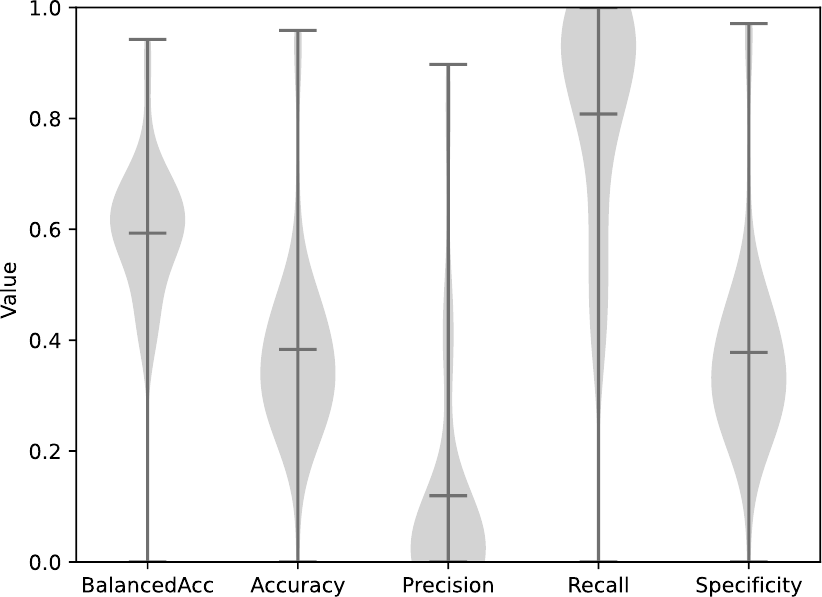}
    }
    \caption{Classification metrics for the five different rule pruning and sorting strategies.}
    \label{fig:rq1-pruning}
\end{figure}

Figure~\ref{fig:rq1-pruning} shows the classification metrics obtained for each rule pruning and sorting strategy. Each violin plot aggregates the results across random seeds and datasets. CPAR has the best accuracy and specificity values, but also results in low recall and balanced accuracy. These metrics are higher when SCBA, CMAR or CBA are applied. In general, it is difficult to decide which strategy works best, as it depends on the metric of interest. However, we believe that SCBA provides the best balance between metrics and only precision is low. Given the imbalanced nature of the paper selection problem, it seems reasonable to focus on balanced accuracy as the most important metric. In this sense, SCBA obtains the best value and therefore it is the chosen strategy to continue with the evaluation of the G3P algorithm.

The following analysis concerns the replacement strategy in the G3P algorithm. Table~\ref{tab:rq1-replacement} shows the results for each dataset and metric, reporting the average and standard deviation. The best value for each metric is highlighted in bold. With the exception of the AISE dataset, NPOP is the best choice in terms of balanced accuracy, recall and specificity. NPOP is also the best in terms of recall for the AISE dataset. We suspect that the different behaviour observed for this dataset could respond to the different proportion of positive instances compared to the other datasets. In any case, the three replacement strategies show quite similar behaviour in terms of balanced accuracy for the AISE dataset. Therefore, we select the NPOP because it presents the best balanced accuracy for four of the five datasets, and the second best value for the AISE dataset.

\begin{sidewaystable}
\centering
\scalebox{0.8}{
\begin{tabular}{lllllll}
\hline
Dataset & Rep. Strategy & Bal. Accuracy & Accuracy & Precision & Recall & Specificity \\
\hline
\multirow{3}{*}{Hall}            & NPOP & \bm{$0.8634\pm 0.0110$}  & \bm{$0.7683\pm 0.0246$} & $0.0535\pm 0.0060$   & \bm{$0.9610\pm 0.0185$} & \bm{$0.7658\pm 0.0251$}  \\
                & ELIT & $0.7509\pm 0.1011$  & $0.593\pm 0.1726$   & \bm{$0.0637\pm 0.0461$}  & $0.9128\pm 0.0715$  & $0.589\pm 0.1746$ \\
                & PGEN & $0.6471\pm 0.1508$  & $0.4564\pm 0.2824$  & $0.0417\pm 0.0247$  & $0.8427\pm 0.1829$  & $0.4514\pm 0.2866$ \\
\hline
\multirow{3}{*}{Kitchenham}      & NPOP & \bm{$0.7188\pm 0.0916$}  & \bm{$0.5996\pm 0.2976$} & $0.0557\pm 0.0340$  & \bm{$0.8447\pm 0.1463$} & \bm{$0.5929\pm 0.3094$}  \\
                & ELIT & $0.6058\pm 0.0471$  & $0.590\pm 0.2028$   & \bm{$0.0791\pm 0.0679$}  & $0.6219\pm 0.1611$  & $0.5896\pm 0.2123$ \\
                & PGEN & $0.609325\pm 0.0236$  & $0.4697\pm 0.0570$  & $0.0586\pm 0.0070$   & $0.7567\pm 0.0482$  & $0.461\pm 0.0594$ \\
\hline
\multirow{3}{*}{Wahono}          & NPOP & \bm{$0.8183\pm 0.0141$}  & \bm{$0.6685\pm 0.0479$} & $0.0264\pm 0.0054$   & \bm{$0.9709\pm 0.0238$} & \bm{$0.6658\pm 0.0485$}  \\
                & ELIT & $0.6859\pm 0.1560$  & $0.5897\pm 0.1932$  & $0.0243\pm 0.0129$   & $0.7838\pm 0.1913$  & $0.5880\pm 0.1944$ \\
                & PGEN & $0.7013\pm 0.1409$  & $0.5522\pm 0.3024$  & \bm{$0.0314\pm 0.0230$}  & $0.8532\pm 0.1222$  & $0.5495\pm 0.3056$   \\
\hline
\multirow{3}{*}{Muthu}           & NPOP & \bm{$0.681\pm 0.0804$} & \bm{$0.8546\pm 0.0550$} & \bm{$0.0954\pm 0.0644$}  & \bm{$0.4964\pm 0.1897$} & \bm{$0.8656\pm 0.0598$}  \\
                & ELIT & $0.5564\pm 0.1119$  & $0.7690\pm 0.1716$  & $0.0456\pm 0.0302$ & $0.3309\pm 0.2029$  & $0.7819\pm 0.1789$ \\
                & PGEN & $0.5419\pm 0.1062$  & $0.7492\pm 0.2168$  & $0.0362\pm 0.0311$ & $0.3226\pm 0.2692$  & $0.7613\pm 0.2282$ \\
\hline
\multirow{3}{*}{AISE}            & NPOP & $0.4982\pm 0.0002$  & $0.2419\pm 0.0907$  & $0.2016\pm 0.0336$ & \bm{$0.9447\pm 0.1573$} & $0.0518\pm 0.1578$ \\
                & ELIT & \bm{$0.4994\pm 0.0274$} & \bm{$0.4576\pm 0.1611$}   & \bm{$0.2816\pm 0.0723$}  & $0.5723\pm 0.2205$ & \bm{$0.4265\pm 0.2636$}  \\
                & PGEN & $0.4966\pm 0.0524$  & $0.4049\pm 0.1727$  & $0.2708\pm 0.0780$  & $0.6563\pm 0.2064$  & $0.3368\pm 0.2711$ \\
\hline
\end{tabular}
}
\caption{Results of classification metrics per dataset and replacement strategy.}
\label{tab:rq1-replacement}
\end{sidewaystable}

Finally, we run our algorithm under different parameter settings (see Table~\ref{tab:parameters}), keeping NPOP as replacement strategy and SCBA for rule pruning and sorting. The results are so similar that we do not report them for the sake of readability.\footnote{For the interested reader, the detailed results are available in the replication package.} We only perceive a slightly higher standard deviation for recall and specificity in the AISE dataset. Due to small influence of the selected parameters, we keep the default values: $maxGen=100$, $popSize=100$, $crossProb=0.9$ and $mutProb=0.1$. To summarise RQ1, we conclude that our algorithm is less sensitive to the evolution parameters than to the strategy for replacing individuals and choosing rules for classifier construction. Both components behave similar for four of the five datasets studied, while the algorithm may perform better with a specific configuration for AISE dataset probably due to a less imbalanced class distribution.

\subsection{RQ2: Influence of bibliometric information}
\label{subsec:RQ2-results}

\begin{figure}[ht]
    \centering
    \subfloat[Hall dataset\label{subfig:rq2-hall}]{%
      \includegraphics[height=4cm]{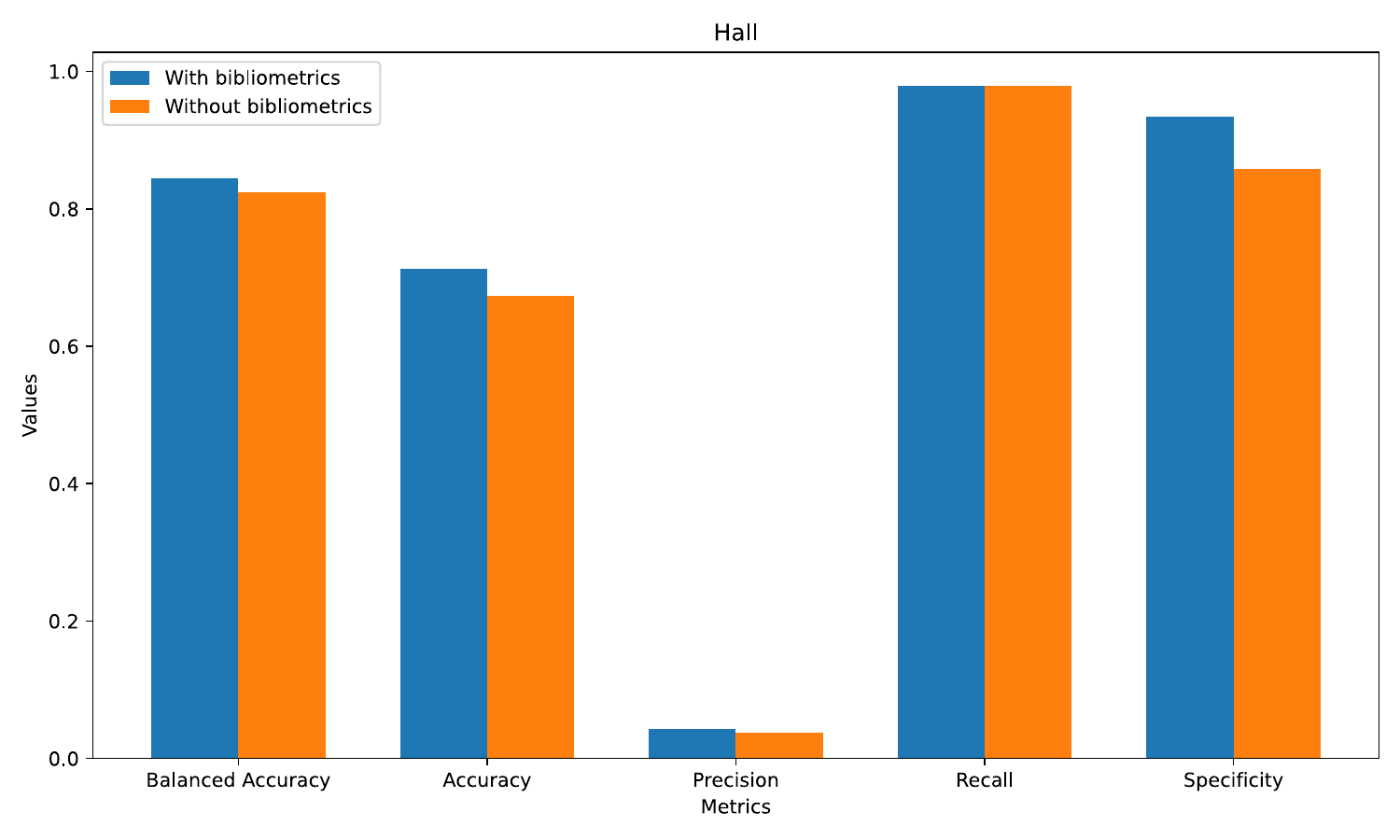}
    }
    \subfloat[Muthu dataset\label{subfig:rq2-muthu}]{%
      \includegraphics[height=4cm]{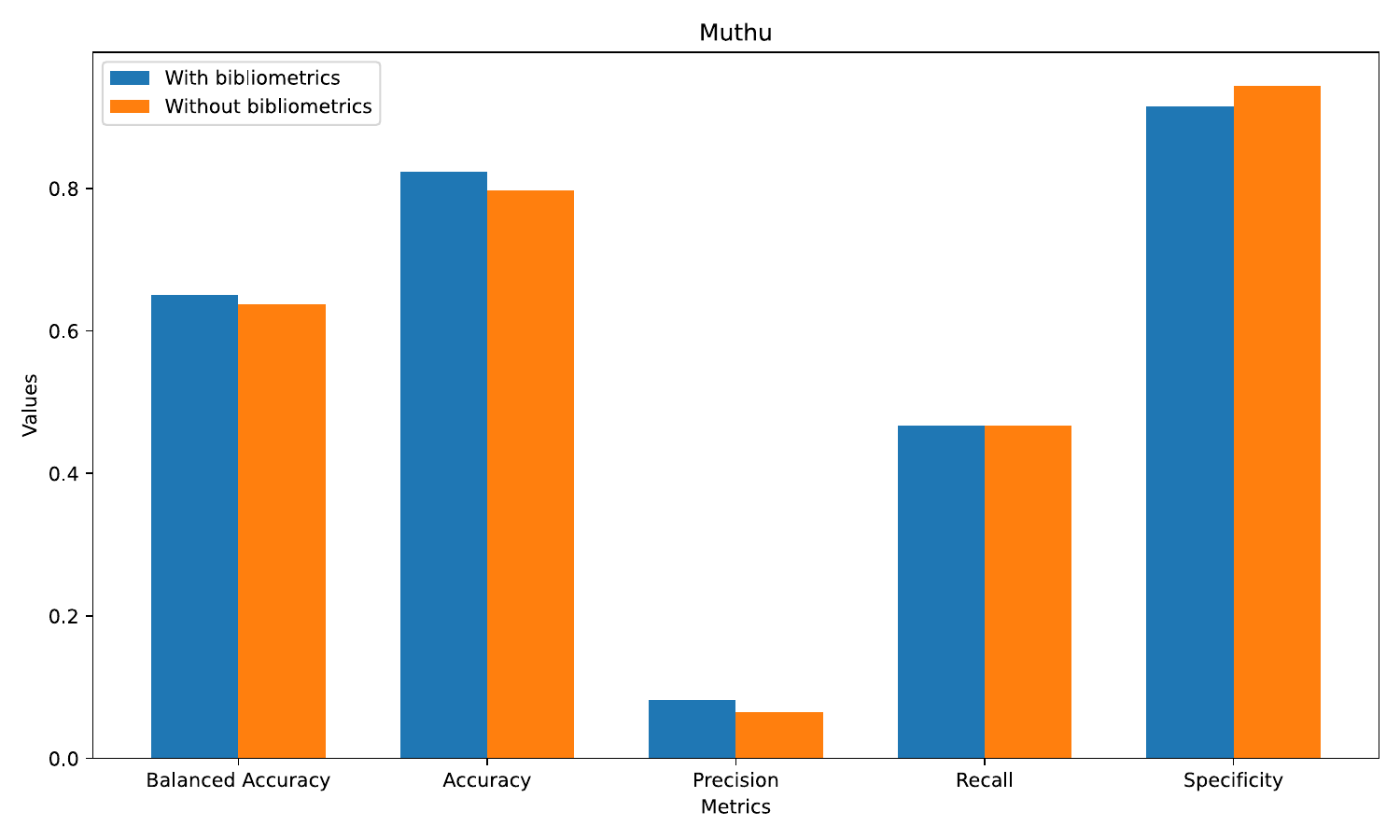}
    }
    \\
    \subfloat[AISE dataset\label{subfig:rq2-aise}]{%
     \includegraphics[height=4cm]{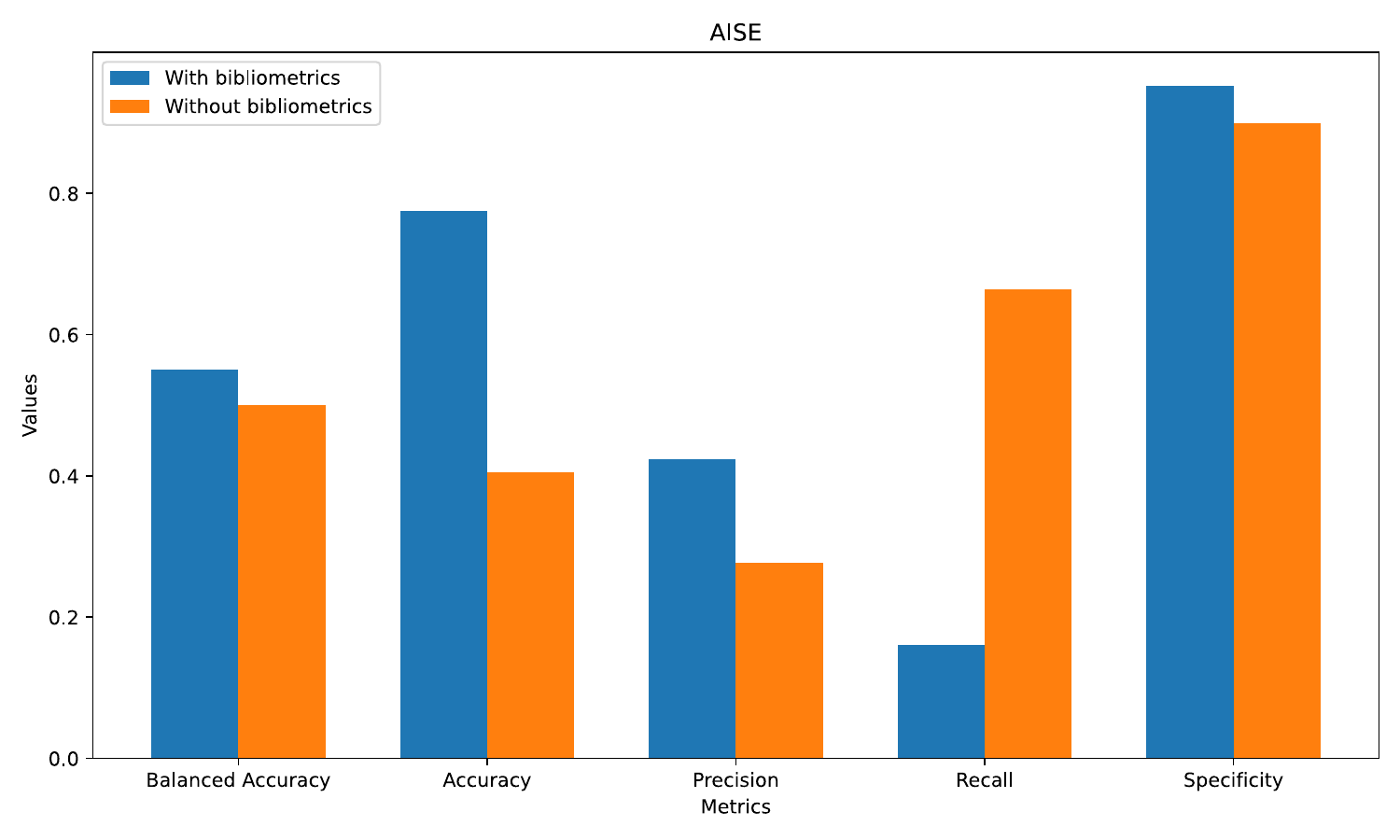}
    }
    \caption{Influence of the bibliometric information in the classification results.}
    \label{fig:rq2-bibliometric}
\end{figure}

Figure~\ref{fig:rq2-bibliometric} shows the results of the classification metrics for two versions of our algorithm, one including bibliometric information and other excluding this information from the CFG. In general, the version using bibliometric information improves the precision, accuracy and recall values. The only exception is the AISE dataset, for which recall suffers a considerable reduction. However, the accuracy and precision improve considerably for this dataset, with increases of 91\% and 53\%, respectively.

We have inspected the best rules obtained in this experiment to analyse the use of bibliometric information. The bibliometric operator that appears most frequently is the number of authors, with 37.5\% of occurrence, followed by the year of publication (31.6\%) and the number of citations (30.89\%). The rules in which these operators appear are usually more specific, \ie the range of values is narrower or the rules contain more comparisons, and they are always combined with text operators. Interestingly, we observe that 92\% of the rules with bibliometric operators that were selected to build classifiers contain the $isCandidate==True$ clause in the consequent. This fact supports our hypothesis that bibliometric operators can greatly benefit the automatic selection of primary studies, as they appear frequently in rules describing positive instances. From the analysis presented, we can conclude that the use of bibliometric information, such as number of citations, number of authors and type of publication, contributes to solve the automatic paper selection problem.

\subsection{RQ3: Interpretability of the rules}
\label{subsec:RQ3-results}

Apart from the use of bibliometric information, the main difference of our algorithm with respect to state-of-the-art proposals is the interpretability of the resulting classifier. The association rules, placed in a correct order, determine how the classifier selects the outcome for a paper (selected or not selected). The rules make the knowledge explicit in a readable format, which makes it easier for the researcher to interpret the classifier's prediction. Table~\ref{table:rules-rq3} summarises some of the properties of the rules used to construct classifiers for the five datasets. The columns represent the mean number of rules composing the final classifier (CR), the average length of the rules found by the G3P algorithm (RL), the total number of rules generated (TR), and the number of bibliometric operators appearing in the rules. The results are presented separately for positive and negative cases.

\begin{table}[]
\scalebox{0.75}{
\begin{tabular}{|l|rrrr|rrrr|}
\hline
\multirow{2}{*}{Dataset} & \multicolumn{4}{c|}{Positive instances} & \multicolumn{4}{c|}{Negative Instances}    \\ \cline{2-9}
            & CR (mean \textpm stdev) & RL    & TR   & BO  & CL (mean \textpm stdev)  & RL & TR & BO \\ \hline
Hall        & 3.2424 \textpm 0.9268 & 7.4474  & 189  & 33  & 3.0404 \textpm 0.9249  & 7.0000 & 178 & 0  \\ \hline
Wahono      & 3.2352 \textpm 0.7445 & 10.8571 & 188  & 51  & 3.8487 \textpm 1.0221  & 7.0000 & 151 & 0  \\ \hline
Kitchenham  & 3.3982 \textpm 1.1537 & 8.3762  & 201  & 52  & 3.8849 \textpm 1.075   & 7.0000 & 150 & 0  \\ \hline
Muthu       & 2.1818 \textpm 0.7472 & 9.8361  & 201  & 20  & 3.0303 \textpm 0.9524  & 7.0000 & 69  & 0  \\ \hline
AISE        & 1.7676 \textpm 0.5683 & 7.4837  & 152  & 12  & 2.9191 \textpm 0.8040  & 7.4257 & 147 & 15 \\ \hline
Total/Avg.  & & 8.8001 & 931 & 168  &  & 7.0851 & 695 & 15 \\ \hline
\end{tabular}
}
\caption{Properties of the classifiers and rules describing positive (selected papers) and negative (non-selected papers) instances used for automatic paper selection.}
\label{table:rules-rq3}
\end{table}

As can be seen, the classifiers are always composed of less than 10 rules, with a similar distribution between rules describing positive and negative instances. It is noticeable that \ourmodel is able to find a large number of rules (931 in total) to describe positive instances despite the high imbalance of some datasets. In fact, this number is larger than the number of rules describing negative instances (695), which shows that our G3P search process guided by a problem-oriented fitness function is suitable. On average, rules describing positive instances are longer in all datasets, especially in the Wahono and Muthu datasets. We suspect that the smaller number of positive instances requires rules with more conditions to correctly distinguish them from the majority of negative samples. Certainly, in the case of the less imbalanced dataset (AISE), the average length of rules describing positive and negative instances is quite similar. As mentioned in Section~\ref{subsec:RQ2-results}, bibliometric operators appear more frequently in the rules describing positive instances. The analysis by dataset clearly corroborates this fact. In four of the five datasets, bibliometric operators appear only in rules describing positive instances. They represent approximately $1/5$ of the rules. Again, \ourmodel shows a different behaviour for the AISE dataset, where a few bibliometric operators are used to generate rules describing negative instances. It should be noted that all cases of bibliometric operators in the Wahono and Kitchenham datasets correspond to the year of publication, as the other operators (number of citations, number of authors and type of publication) are not available for these datasets.

\begin{figure}[ht]
    \centering
    \includegraphics[width=\linewidth]{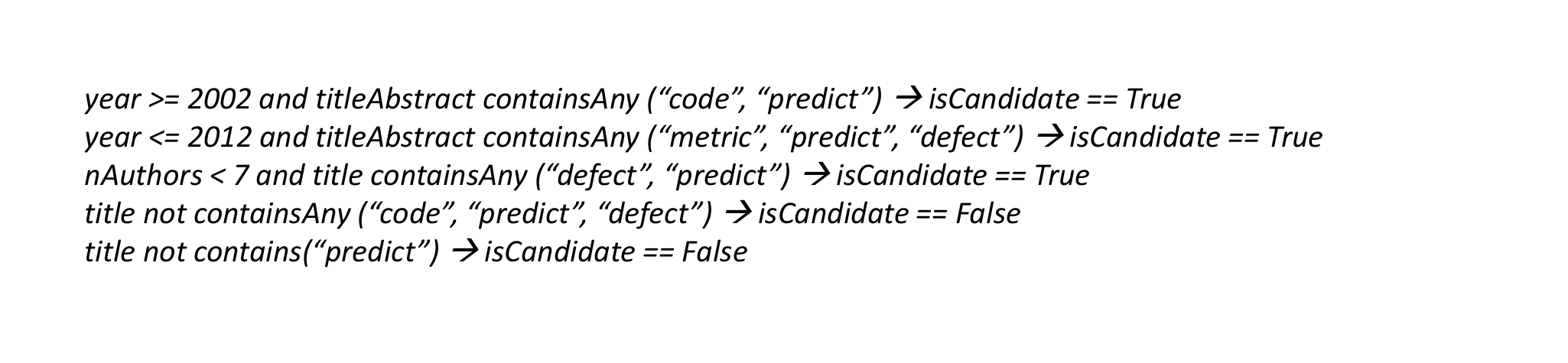}
    \caption{Sample rules output for Hall dataset}
    \label{fig:rq3-rules}
\end{figure}

As an illustrative example, we analyse a rule-based classifier generated for the Hall dataset. Figure~\ref{fig:rq3-rules} shows the sorted list of rules that compose the classifier. The first three rules specify the conditions that candidate papers selected for the SLR must satisfy. The last two rules cover negative instances, \ie papers that do not fit the SLR topic. In this case, the rules tend to be shorter and more general to cover the remaining instances not classified by the previous rules. The rules obtained are easy to understand and combine both textual and bibliometric information. For instance, the first rule in Figure~\ref{fig:rq3-rules} suggests that papers published after 2002 and whose title or abstract contains terms like ``code'' and ``predict'' are good candidates for primary studies. The second and third rules pose additional restrictions on the publication year, the number of authors and the terms appearing in title and abstract. Complementarily, the classifier has detected that papers that do not use the terms ``code'', ``predict'' or ``defect'' in the title are not representative of the SLR topic (defect prediction in software engineering).

In response to RQ3, we believe that rules such as these can help a researcher understands what characteristics of the research papers make them potential primary studies. The flexibility of the CFG configuration also allows researchers to be more or less specific in the keywords and metadata information they consider relevant for paper selection. In this sense, the possibilities of defining a specific vocabulary and limiting the range of numerical conditions (citations, authors and years) enrich the automatic paper selection process.

\subsection{RQ4: Comparison with the state-of-the-art}
\label{subsec:RQ4-results}

Table~\ref{tab:rq4-results} shows the results of the classification metrics obtained after running \ourmodel with its general-purpose configuration and FAST2 following the methodology described in Section~\ref{subsec:methodology}. For \ourmodel, we report the mean and standard deviation. The best value for each metric is highlighted in bold. We also provide the results of the statistical tests, compiling the results from the five datasets. As can be seen, our approach obtains better balanced accuracy for all datasets. The improvement is considerable for the Wahono and Muthu datasets. Low precision values are common due to the imbalanced nature of the classification problem, but we still provide better results than FAST2 for all datasets except AISE. According to the Wilcoxon test, \ourmodel is superior to FAST2 in terms of recall. 

We also discuss the execution time required by each algorithm. The number of instances in the dataset appears to be the most influential factor. For the largest dataset (AISE), our algorithm took an average of $79.4\pm3.5$ minutes to complete the run with all folds. In contrast, FAST2 took 5 days to complete the learning process. Inspecting the log messages from FAST2, we observe that its performance decreases over time. Its simulation process labels the studies in batches of ten instances, and then retrains the SVM classifier. If a large dataset is provided, this process becomes quite expensive, as each batch iteration involves revisiting the entire set of labelled papers. 

To conclude our analysis of RQ4, we can state that, in the comparison scenario, our algorithm provides better performance than FAST2. Not only do we obtain higher balanced accuracy, but we also complete the learning process without human intervention in less time for large datasets, as is usually the case for SLR automatic search results.

\begin{sidewaystable}[!ht]
\scalebox{0.65}{
\begin{tabular}{lllllll}
\hline
Dataset    & Algorithm & Bal. Accuracy & Accuracy & Precision & Recall & Specifity \\ \hline
\multirow{2}{*}{Hall}  & \ourmodel  & \textbf{0.8712\textpm0.0039} & \textbf{0.7868\textpm0.0091} & \textbf{0.0594\textpm0.0023} & \textbf{0.9579\textpm0.0026}  & \textbf{0.7846\textpm0.0093 } \\ 
      & FAST2 & 0.7308 & 0.7830     & 0.0355     & 0.6774   & 0.7842      \\ \hline
\multirow{2}{*}{Kitchenham} & \ourmodel  & \textbf{0.7352\textpm0.0063} & \textbf{0.6441\textpm0.0171} & \textbf{0.0610\textpm0.0024} & \textbf{0.8315\textpm0.0202}  & \textbf{0.639\textpm0.0180}  \\ 
      & FAST2 & 0.6478 & 0.5175      & 0.0431      & 0.7857   & 0.5100      \\ \hline
\multirow{2}{*}{Wahono}     & \ourmodel  & \textbf{0.8199\textpm0.0029} & 0.6795\textpm0.0079  & \textbf{0.0267\textpm0.0006} & \textbf{0.9629\textpm0.0034}       & 0.6769\textpm0.0080   \\ 
      & FAST2 & 0.6672 & \textbf{0.7539}     & 0.0211     & 0.5789   & \textbf{0.7555}      \\ \hline
\multirow{2}{*}{Muthu}      & \ourmodel  & \textbf{0.7034\textpm0.0233} & 0.7977\textpm0.0283 & \textbf{0.0851\textpm0.01466} & \textbf{0.6035\textpm0.0768}       & 0.8033\textpm0.0314  \\ 
      & FAST2 & 0.5  & \textbf{0.9708}     & 0.0000    & 0.0000 & \textbf{1.0000}        \\ \hline
\multirow{2}{*}{AISE}       & \ourmodel  & \textbf{0.4982\textpm0.0002} & \textbf{0.2420\textpm0.0670} & 0.2017\textpm0.0248    & \textbf{0.9445\textpm0.1162}     & 0.0519\textpm0.1166  \\ 
      & FAST2 & 0.4307 & 0.2297     & \textbf{0.7114}       & 0.3548  & \textbf{0.3473}      \\ \hline
\multicolumn{2}{l}{\textit{Wilcoxon test} (p-value)}   & \textit{Not rejected}
(0.1171) & \textit{Not rejected} (0.9168) & \textit{Not rejected} (0.2506) & \textit{Rejected} (0.0282) & \textit{Not rejected} (0.9168)\\
\hline
\end{tabular}
}
\caption{Comparison between our proposal (\ourmodel) and FAST2 in terms of classification metrics.}
\label{tab:rq4-results}
\end{sidewaystable}

\section{Concluding remarks}
\label{sec:conclusions}

This paper presents \ourmodel, an evolutionary machine learning approach to support primary study selection in systematic literature reviews. \ourmodel is designed to be highly flexible, as it provides researchers with specific configurable components to decide the textual and bibliometric information retrieved from papers that would serve to make automatic decisions. Based on the use of a context-free grammar, \ourmodel builds an interpretable rule-based classifier by adopting an intelligent exploration of the search space. The search process, which uses G3P and a problem-specific fitness function to deal with data imbalance, is able to generate comprehensible rules that inform the researcher about the most important properties for selecting relevant scientific papers from the results of a literature search.

\ourmodel has been evaluated from several perspectives beyond classification performance, using five different datasets derived from SLR studies. Our first experiment analyses how different configurations of the internal search components affect \ourmodel. The results show some differences depending on the dataset, although we arrive at a general-purpose configuration with competitive results on the majority of performance metrics. Our second experiment corroborates the contribution of bibliometric information to the problem of automatic paper selection, a type of information not previously considered by any proposal in the literature. Bibliometric information has proven to be very useful in generating specific rules to identify relevant papers, which are more difficult to classify correctly due to the imbalanced nature of SLR datasets. From the interpretability point of view, we analyse the obtained classifiers in terms of number of rules and rule length. Finally, \ourmodel provides better results than FAST2, a black-box active learning classifier.

Future work will focus on defining additional bibliometric operators and implementing more specific G3P components, \ie genetic operators, to exploit such information. Researcher involvement could also be encouraged in different ways. For example, the design of an interactive evolutionary algorithm would allow the redefinition of the vocabulary during the search, as well as the dynamic adjustment of the grammar operators based on the SLR topic. Although the code for \ourmodel is currently available, it would be interesting to offer a web-based solution so that researchers could upload the results of their SLR searches and adapt the execution to their needs. 
The emergence of LLMs to assist researchers in paper selection and other SLR tasks also requires further attention. LLMs could be useful to extract more semantic information, while supervised white-box methods could act as a peer reviewer to ensure more accurate and transparent decisions.

\section*{Supplementary material}
As supplementary material, we provide the source code of \ourmodel. Detailed results per dataset and configuration, as well as statistical analysis, are also available. This supplementary material can be accessed from the following Zenodo repository: \url{https://doi.org/10.5281/zenodo.17218905}.

\section*{Acknowledgement}
\emph{Funding:} This work has been supported by the Spanish Ministry of Science, Innovation and Universities, grant PID2023-148396NB-I00 funded by MICIU/AEI /10.13039/501100011033.







\end{document}